\documentclass[sigconf, language=english]{acmart}
\AtBeginDocument{%
  }
\usepackage{multirow} 
\usepackage{xeCJK}
\usepackage{balance}
\usepackage{xcolor}
\usepackage[most]{tcolorbox} %
\usepackage{listings}
\usepackage{makecell} %

\setcopyright{acmlicensed}
\copyrightyear{2026}
\acmYear{2026}
\setcopyright{cc}
\setcctype{by-nc-nd}
\acmConference[ICAIL 2026]{21st International Conference on Artificial Intelligence and Law}{June 08--12, 2026}{Singapore, Singapore}
\acmBooktitle{21st International Conference on Artificial Intelligence and Law (ICAIL 2026), June 08--12, 2026, Singapore, Singapore}
\acmDOI{10.1145/3836937.3836950}
\acmISBN{979-8-4007-2756-6/2026/06}

\begin{document}

\title{TW-LegalBench: Measuring Taiwanese Legal Understanding}

\author{Fei-Yueh Chen}
\email{fchen27@ur.rochester.edu}
\affiliation{
    \institution{University of Rochester}
    \city{Rochester}
    \state{NY}
    \country{USA}
}

\author{Chun Huang Lin}
\email{huang0jin@gmail.com}
\affiliation{
    \institution{National Taiwan University}
    \city{Taipei}
    \country{Taiwan}
}
\author{Chan Wei Hsu}
\email{hcw14002@gmail.com}
\affiliation{
    \institution{National Taiwan University}
    \city{Taipei}
    \country{Taiwan}
}
\author{Kuan Hsuan Yeh}
\email{graceyeh200@gmail.com}
\affiliation{
    \institution{National Taiwan University}
    \city{Taipei}
    \country{Taiwan}
}
\author{Zih-Ching Chen}
\email{virginiac@nvidia.com}
\affiliation{
    \institution{NVIDIA}
    \city{Santa Clara}
    \state{CA}
    \country{USA}
}
\author{Kuan-Ming Chen}
\email{kuanmingchen@ntu.edu.tw}
\affiliation{
    \institution{National Taiwan University}
    \city{Taipei}
    \country{Taiwan}
}
\author{Patrick Chung-Chia Huang}
\email{pcchuang@ntu.edu.tw}
\affiliation{
    \institution{National Taiwan University}
    \city{Taipei}
    \country{Taiwan}
}

\renewcommand{\shortauthors}{Chen, et al.}

\begin{abstract}
Large language models (LLMs) have shown impressive capabilities across diverse tasks, yet their performance on jurisdiction-specific legal reasoning remains underexplored. We present TW-LegalBench that utilizes Taiwanese legal system's rich official corpus open to the public to fill the gap in evaluating LLMs on Taiwanese law, among common-law benchmarks that focus on English sources and civil-law benchmarks focusing on sources of Simplified Chinese. TW-LegalBench comprises three task types: (1) over 16,000 multiple-choice questions (MCQs) across five years of official examinations in 18 professional domains; (2) 117 open-ended essay questions (OEQs) from examinations for legal professionals with official scoring rubrics; and (3) more than 14,000 legal judgment prediction (LJP) instances covering hundreds of crime categories. We evaluate 13 LLMs using accuracy for MCQs, a decomposed LLM-as-Judge framework based on the scoring rubric points for OEQs, and metrics for sentencing accuracy and statute citation for LJP. Our results reveal that top-performing models exceed the passing threshold for qualified lawyers (passing rate: 11\%) but fall short of that for judges and prosecutors (passing rate: 1$\sim$2\%). For LJP, while models demonstrate reasonable verdict type accuracy and sentence prediction capability, they struggle to cite exact legal articles. These findings highlight that reliable legal text generation remains challenging for LLMs, even though their performance on qualification examinations approaches human level.
\end{abstract}

\ccsdesc[500]{Applied computing~Law}
\ccsdesc[500]{Computing methodologies~Reasoning about belief and knowledge}
\ccsdesc[500]{Information systems~Evaluation of retrieval results}
\ccsdesc[500]{Information systems~Question answering}
\ccsdesc[300]{Information systems~Specialized information retrieval}

\keywords{Generative AI, Large Language Models, Benchmark,  Traditional Chinese, Legal Exam, Verdict Prediction}
\begin{teaserfigure}
  \includegraphics[width=\textwidth]{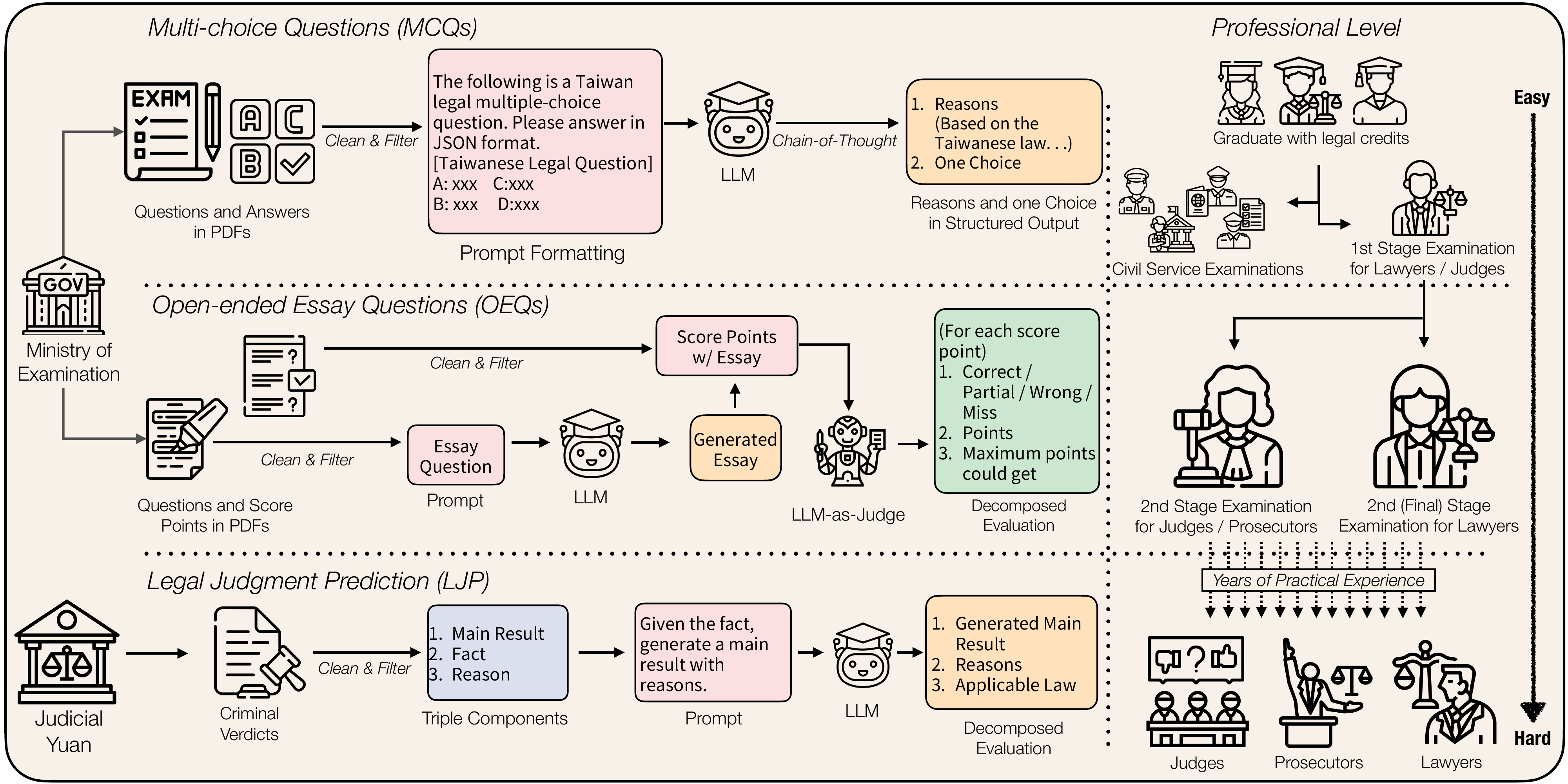}
  \caption{Main Framework for TW-LegalBench}
  \Description{}
  \label{fig:teaser}
\end{teaserfigure}

\maketitle

\section{Introduction}

Large language models (LLMs), including model families such as LLaMA \cite{llama3herd}, Qwen \cite{Yang2024Qwen25TR, yang2025qwen3technicalreport}, and GPT \cite{OpenAI_GPT4_2023, OpenAI2025GPT5}, are developed either as open models for broad reuse or as proprietary systems that can be adapted to specific domains via fine-tuning. Among high-impact application areas, law is a particularly promising yet challenging domain for text generation. On the one hand, LLMs are expected to promote access to justice for those who cannot afford legal services. On the other hand, tasks such as legal reasoning require precise interpretation, structured reasoning, and sensitivity to jurisdiction-specific doctrines. As a result, a benchmark that evaluates LLMs' performance on such tasks is important in both technical and policy aspects.

English benchmarks such as MMLU \cite{hendryckstest2021} and LegalBench \cite{10.5555/3666122.3668037} include tasks about the common law system. For benchmarks related to Taiwanese culture in Traditional Chinese, TMLU \cite{DBLP:journals/corr/abs-2403-20180} includes lawyer-exam multiple-choice questions for one year, and TMMLU+ \cite{tam2024tmmlu} collects several standardized tests covering auditing and finance law. Prior work has also proposed legal benchmarks for other jurisdictions and languages, such as LegalBench \cite{10.5555/3666122.3668037} (common-law tasks), LawBench \cite{fei-etal-2024-lawbench}, and LawShift \cite{han2025lawshift} for Simplified Chinese models (civil-law tasks).

Despite the recent progress, current legal benchmarks suffer from three systemic limitations that may obscure a model's true professional readiness. First, many benchmarks \cite{DBLP:journals/corr/abs-2403-20180, hendryckstest2021, tam2024tmmlu} exhibit \textbf{excessive coarseness} by aggregating disparate legal doctrines into overly broad categories. For instance, TMMLU+ \cite{tam2024tmmlu} conflates the Administrative Litigation Act and the Public Functionaries Discipline Act under a single administrative-law category, limiting its ability to provide precise and tailored answers. Second, there is a clear \textbf{assessment gap} due to an over-reliance on closed-ended questions in most law-related benchmarks. While convenient for scoring, these formats fail to simulate real-world problem-solving, which often involves open-ended, customized fact patterns. Finally, a significant \textbf{jurisdictional imbalance} persists. Most benchmarks are designed based on materials from common-law and Anglophone jurisdictions. This leaves civil-law jurisdictions like Taiwan without the large-scale, high-fidelity resources necessary to test alignment with localized legal doctrines.

To address these limitations, we introduce a multi-tiered benchmark grounded in the unique, open-data ecosystem of Taiwan's civil-law tradition. Specifically, we curate our tasks from official examination questions and court verdicts, representing qualification milestones widely recognized across jurisdictions and distinct levels of legal service: rule clarification and recitation, hypothetical problem-solving, and real-world dispute resolution.
First, we establish a foundation of rule-based recognition through 16,493 statute-level annotated questions, allowing for a high-resolution diagnosis of model expertise across 43 distinct law types. Second, we challenge models to perform complex synthesis via 117 open-ended bar-exam essay questions, utilizing a validated LLM-as-judge framework to assess the construction of coherent legal arguments. Finally, we evaluate real-world alignment through a large-scale corpus of 14,325 criminal judgments, tasking models to predict judicial outcomes across 107 crime categories. In sum, our contributions are as follows:

\begin{itemize}
  \item \textbf{Statute-level annotation for MCQs.} We collect official legal examinations across five years (2020--2024), covering 16,493 multiple-choice questions across 18 professional domains, including civil service, judicial, and professional examinations. We manually annotate each question with its relevant statute (e.g., the Civil Code, the Criminal Code, etc.), identifying 43 distinct law types across 6 legal categories. This enables finer-grained capability analyses than previous benchmarks.
  
  \item \textbf{An LLM-as-Judge evaluation framework for open-ended tasks.} We establish a rigorous evaluation protocol for open-ended legal reasoning, employing an LLM-as-judge framework calibrated against official scoring rubrics.
  
  \item \textbf{A large-scale verdict prediction dataset.} We curate 14,325 criminal judgments from 2013 to 2024, covering 107 crime categories after conservative normalization. We design a balanced evaluation split with 5 samples per crime type (535 test cases), ensuring fair assessment across all categories regardless of their frequency in real-world data.

\end{itemize}

\section{Related Work}

Our benchmark aims to fill the gap in evaluating models' legal reasoning capacity in a non-English-speaking jurisdiction that follows the civil-law tradition.
Current benchmarks either focus on English-speaking jurisdictions that follow the common-law tradition or, even in a civil-law context, lack a comprehensive scope of tasks (Table~\ref{tab:comparison}).

\subsection{Benchmarks Based on Common-Law Jurisdictions}
Early benchmarks designed to evaluate multiple dimensions of LLMs were predominantly English-based, resulting in these benchmarks' emphasis on common-law systems. Questions from subjects that are dominated by common-law rules focus on comparison with precedents and unearthing the implicit rules therefrom. This can be observed in the MMLU \cite{hendryckstest2021} bar examination task, in which questions involving Torts in the Professional Law section present hypothetical scenarios and require models to compare the current scenarios with previous cases to determine the applicable rules and predict the most likely outcome. Likewise, the Rule-application category (one of the six reasoning types) in LegalBench \cite{10.5555/3666122.3668037} encompasses a diverse array of cases through which models are expected to perform sound legal reasoning. In short, benchmarks utilizing data from common-law jurisdictions seem to emphasize comparisons of cases and precedents. Admittedly, benchmarks from common-law jurisdictions also have to deal with statutory interpretation, such as calculating tax owed according to tax law in LegalBench \cite{10.5555/3666122.3668037}; however, such tasks account for a relatively smaller and less frequent subset. Given that these benchmarks are fundamentally grounded in common-law systems and Anglo-American culture, standards such as MMLU may not be easily generalizable to non-English-speaking jurisdictions \cite{singh-etal-2025-global}.

\subsection{Benchmarks Based on Civil-Law Jurisdictions}
In civil-law systems, rules are explicitly prescribed in the form of abstract statutes. Instead of comparing the fact pattern of the case at hand with precedents, judges in civil-law courts directly apply the rules to the case. In this sense, legal reasoning in the civil-law tradition demonstrates a different pattern from that in the common-law tradition. Namely, comparison between the current case and precedents is less relevant, at least in examinations. Moreover, legal reasoning in civil-law jurisdictions is sensitive to statutory wording. This has been observed in LawBench \cite{fei-etal-2024-lawbench}, a benchmark based on the Chinese (civil-law) legal system. It evaluates the capability of reciting statutes and predicting judgments under two conditions: with and without the provision of relevant statutory texts. Their results show that models generally perform better when provided with relevant texts, demonstrating that models may find it challenging to recite correct statutes. Moreover, legal reasoning in civil-law systems relies not only on statutes \textit{per se} but also on their context, such as legislative history and amendments. Changes in statutory wording can play a significant role in statutory interpretation. This presents a significant challenge for current LLMs, whose training data may lack comprehensive coverage of statutory revision histories. Models may also struggle to recognize the differences between pre- and post-amendment versions of laws—or fail to treat successive versions as part of a coherent legal evolution. LawShift \cite{han2025lawshift} is the first benchmark specifically designed to evaluate model performance on statutory amendments, using the historical evolution of Chinese Criminal Codes as its foundation. The authors found that state-of-the-art models are remarkably fragile: their reasoning fails to reflect the changes brought by newly enacted provisions and instead defaults to outdated legal concepts embedded in their original training data.

\subsection{Traditional Chinese and Taiwanese Law}

To address these limitations, we utilize data from the Taiwanese legal system, which follows the civil-law tradition under strong German and Japanese influences. Although the Taiwanese government has published rich and high-quality legal data, to our knowledge, no benchmark has been developed specifically for evaluating models on Taiwanese law. TMLU \cite{DBLP:journals/corr/abs-2403-20180} compiled a limited number of legal examination questions of varying difficulty levels, including hundreds of multiple-choice questions extracted from a series of national examinations, such as the bar examination, driver's license tests, and junior and senior high school civics examinations. TMMLU+ \cite{tam2024tmmlu} includes some audit-related questions and general law items, yet these administrative law questions primarily assess fundamental legal \textit{knowledge} rather than legal \textit{reasoning} derived from statutory provisions and case analysis discussed above. Nevertheless, questions included in TMLU and TMMLU+ are limited in scope and often too abstract to be informative in specific cases. To the best of our knowledge, LLAWA \cite{chen2025continual} represents the only existing study that focuses on Traditional Chinese, incorporates legal corpora for fine-tuning, and targets legal reasoning tasks. Its tasks are divided into multiple-choice questions from the bar examination and the Taiwan Jurist Journal, essay questions on criminal law from the bar examination, and legal reasoning problems drawn from judicial symposia. The primary contribution of LLAWA lies in its exploration of various fine-tuning methods to enhance model reasoning capabilities, along with a discussion of the trade-off between performance on multiple-choice and essay questions. However, the study does not further examine model performance at specific steps within the legal reasoning process or analyze the patterns of errors that arise. By evaluating different training strategies solely through final scores, it remains difficult to determine whether models genuinely comprehend legal logic or merely engage in pattern matching based on memorized corpora.

\begin{table}[h]
  \caption{A comparison of Taiwanese legal benchmarks. Only professional questions are counted. OEQs and LJP are counted as Open Questions.}
  \label{tab:comparison}
  \begin{tabular}{cccc}
    \toprule
    Benchmark & MCQs & Open Questions & Total \\
    \midrule
    TMLU \cite{DBLP:journals/corr/abs-2403-20180} & 279 & 0 & 279\\
    TMMLU+ \cite{tam2024tmmlu} & 1,890 & 0 & 1,890\\
    LLAWA \cite{chen2025continual} & 904 & 1,894 & 2,798 \\
    TW-LegalBench (ours) & 16,493 & 117 + 14,325 & 30,935 \\
    \bottomrule
  \end{tabular}
\end{table}

\section{Dataset}

TW-LegalBench consists of three main components: \\Multiple-choice Questions (MCQs), Open-ended Essay Questions (OEQs), and Legal Judgment Prediction (LJP).

\subsection{Data Sources}

We collect data from two main sources:

\textbf{Ministry of Examination, Taiwan.} We obtain MCQs and OEQs from the official website of the Ministry of Examination.\footnote{\url{https://wwwc.moex.gov.tw}} This website publishes examination questions and answers for all national examinations in Taiwan, including the bar exam, civil service exams, and professional qualification exams. 

\textbf{Judicial Yuan Law and Regulations Retrieving System.} We collect criminal court judgments from the opendata platform from the Judicial Yuan, Taiwan.\footnote{\url{https://opendata.judicial.gov.tw}} These judgments are publicly available and have been anonymized to protect personal information.

\subsection{Multiple-Choice Questions (MCQs)}
For the MCQs, we first downloaded examination questions and answer keys from the Ministry of Examination, Taiwan. We retained only single-answer legal questions, excluding items with multiple officially accepted correct answers and questions that combined legal knowledge with other subjects on the same examination paper (e.g., legal knowledge combined with English). We collected all officially administered professional legal examinations from 2020 to 2024, yielding a total of 360 files encompassing 85 distinct examination subjects across 18 professional domains. These examinations fall into four major categories: Civil Service, Judicial Personnel, Police Personnel, and Professional and Technical Personnel. It should be noted that the examination subjects vary from year to year, as not all examinations are administered annually, and some entry-level examinations have been consolidated. Table~\ref{tab:questions-year} shows the data distribution by year.

\begin{figure}[h]
\begin{tcolorbox}[colframe=black, colback=white, sharp corners]
\small 
\textbf{Question:} \\
甲已喪偶，有乙、丙二子。乙留學時甲給與新臺幣（下同）150萬元，丙開店營業時甲亦給與150萬元。不久甲死亡，留有現金300萬元。遺產分割時，繼承人乙得分得多少金額之遺產？\\
\textcolor{teal}{(A is predeceased by his/her spouse and has two sons, B and C. A gave B NTD\$1.5 million (same currency below) to support B's pursuit of studies abroad. A also gave C \$1.5 million to support C's newly opened business. Shortly after, A passed away, leaving NTD\$3 million in cash. During the division of the estate, how much inheritance is heir B entitled to receive?)} \\
\textbf{Choices:} \\
\hspace*{3mm}
A: 75萬元 \hspace{5mm} \textcolor{teal}{(NTD\$0.75 million)} \\
\hspace*{3mm}
B: 150萬元 \hspace{3.5mm} \textcolor{teal}{(NTD\$1.5 million)} \\
\hspace*{3mm}
C: 225萬元 \hspace{3.5mm} \textcolor{teal}{(NTD\$2.25 million)} \\
\hspace*{3mm}
D: 250萬元 \hspace{3.5mm} \textcolor{teal}{(NTD\$2.5 million)}\\
\textbf{Answer: C} \\
\vspace{1mm}
\end{tcolorbox}
\caption{An error analysis example from the Civil Code on MCQs where all evaluated models failed to predict the correct answer. 11 out of 13 models chose B, while the remaining two chose A.}
\label{fig:civil-law-failure}
\end{figure}

Following initial data cleaning, we recruited two research assistants\footnote{They are one undergraduate senior and one first-year master's student from a law school in Taiwan.} to annotate each multiple-choice question. The annotators were asked to identify: (1) whether the question pertains to domestic or international law (notably, we did not identify any question that simultaneously involved both domestic and international law), and (2) the specific legal source (domestic statute or international convention) to which the question corresponds. Among all multiple-choice questions, 92.3\% concerned domestic law, 0.8\% concerned international law and the rest of the questions do not concern a specific law (e.g. basic concepts of law or judicial practice). In total, we annotated 781 distinct domestic laws and 46 international laws or conventions; however, 534 of these legal provisions appeared only one to four times in the dataset.

To investigate whether language affects generation quality, we translated all questions into Simplified Chinese and English using Claude-Sonnet-4.5. For the English translations, the prompt incorporated official translation from the Judicial Yuan to ensure terminological consistency.

\begin{table}[h]
  \caption{Distribution of MCQs and OEQs by Year.}
  \label{tab:questions-year}
  \begin{tabular}{ccc}
    \toprule
    Year & MCQs & OEQs \\
    \midrule
      2020 & 3,604 & 24\\  
      2021 & 3,331 & 24\\   
      2022 & 3,433 & 23\\   
      2023 & 3,398 & 23\\ 
      2024 & 2,727 & 23\\  
    \midrule
    Total & 16,493 & 117\\ 
    \bottomrule
  \end{tabular}
\end{table}

Regarding the human baseline, the Ministry of Examination currently publishes only the final weighted admission scores for each professional category examination. These composite scores incorporate results from English, essay writing, and other professional subjects, with legal examinations constituting only a small proportion of the overall score. Consequently, we were unable to obtain subject-specific human performance scores for individual legal topics. If the admission cutoff scores are directly averaged, the approximate accuracy required would be in the range of 70–80\%.

\subsection{Open-Ended Essay Questions (OEQs)}

Among all the professions represented in our collected data, only judges, prosecutors, and lawyers are required to complete essay questions (the second stage of the Judicial and Bar Examination), while other professions are assessed solely by multiple-choice questions or additional interviews. We similarly downloaded the original exam questions from the Ministry of Examination; however, unlike the multiple-choice questions with concrete answer keys, the essay questions are only provided with the scoring rubrics outlining the key points for evaluation. Exam subjects remain stable from 2020 to 2024, with the only exception that the number of Tax Law questions decreased from three to two in 2022.
Table~\ref{tab:questions-year} shows the distribution of MCQs and OEQs by year. 

For the human baseline, we collected scores and related performance statistics from test takers from 2021 to 2024. Data from 2020 for essay questions were excluded because the statistics differed from the current system.

\subsection{Legal Judgment Prediction (LJP)}
For the judgment prediction task, we selected first-instance criminal court judgments from 2013 to 2024 as our source data. We applied the following filtering criteria. First, we retained only judgments with a clearly delineated structure comprising "Main Result," "Facts," and "Reasoning" sections, as certain judgment types (e.g., summary judgments) may consolidate the facts and reasoning into a single section. Second, we established minimum character thresholds for each section of the judgments. After applying stratified sampling, we obtained 14,325 judgments covering more than 700 offense categories.\footnote{The offense categories used here are based on the labels provided by the Judicial Yuan in the original data, rather than the specific criminal statutes or statutory offense names.} We then selected the 107 offense categories that contained at least 10 cases each, and for each category, we sampled 5 judgments for the test set. After this processing, we obtained 13,790 judgments for training data and 535 judgments for test data.

\section{Experimental Setup}
We evaluate the zero-shot chain-of-thought (CoT) \cite{10.5555/3600270.3602070} capabilities of four closed-source models and nine open-source models across all three tasks in TW-LegalBench. Our model selection strategy aims to cover three distinct categories of LLMs. First, we include closed-source models (claude-sonnet-4.5 (claude-sonnet-4-5-20250929), gpt-5 (gpt-5-2025-08-07), gpt-5.2 (gpt-5.2-2025-12-11), and gpt-4o (gpt-4o-2024-08-06)) that represent the current frontier in large language model capabilities. Second, we select open-source models optimized for reasoning tasks, ranging from large-scale systems such as qwen3-235b \cite{yang2025qwen3technicalreport} and llama-3.1-405b \cite{llama3herd} to more compact variants including qwen2.5-7b \cite{Yang2024Qwen25TR}, gpt-oss-120b, gpt-oss-20b \cite{openai2025gptoss120bgptoss20bmodel}, and nemotron-3-nano-30b-a3b \cite{nvidia2025nemotron3nanoopen}. Third, to evaluate the importance of language-specific pretraining, we include models trained primarily on Traditional Chinese data, including llama-taiwan-70b-instruct, llama-taiwan-8b-128k \cite{lin2023taiwanllmbridginglinguistic}, \\and breeze-7b-instruct \cite{hsu2024breeze7btechnicalreport}. 

It is worth noting that, due to the release date, these Traditional Chinese models are built upon earlier architectures such as Llama3 \cite{llama3herd} and Mistral \cite{jiang2023mistral7b}, and prioritize the incorporation of Traditional Chinese and Taiwan-specific cultural data during pretraining or supervised fine-tuning (SFT) stage, in contrast to newer models that emphasize reasoning capabilities through reinforcement learning. All open-source models were accessed via the NVIDIA NIM API, except for the llama-taiwan-8b-128k.

Regarding experimental configurations, we set the temperature to 0 for all models except gpt-5 and gpt-5.2, and applied greedy decoding where supported. For gpt-5 and gpt-5.2, the temperature was set to 1, as these models do not permit a temperature value of 0.

\subsection{Prompting Strategy}
We designed distinct system prompts and user prompts for each of the three tasks, requiring all models to generate outputs in JSON format. For the MCQs involving different language settings, we translated the prompt templates accordingly to ensure linguistic consistency throughout the entire prompt. If a model failed to produce output in the required format, the response was marked as an incorrect answer.
\subsection{Evaluation Metrics}
\subsubsection{MCQs}

Since we excluded questions with multiple correct answers and ensured that each question contains four options with exactly one correct answer among A, B, C, and D, accuracy can be computed directly. Beyond overall accuracy, we analyze performance across multiple dimensions including examination year (2020--2024), examination type (84 distinct types), legal category, and language.

\subsubsection{OEQs}

We adopted an LLM-as-Judge approach, decomposing the overall evaluation into multiple components and outputting discrete results to avoid the instability associated with directly generating numerical scores for the whole question \cite{GU2026101253}. To mitigate potential self-favoritism bias, we use two judge models—gpt-5 and claude-sonnet-4.5—and report the mean score across both judges. First, since each question corresponds to one or more scoring rubric points, and each rubric point maps to specific steps in the answer, we instructed the judge model to output a discrete label and corresponding score for each rubric point. Not all rubric points are accompanied by explicit point allocations. In such cases, we instructed the judge model to distribute the total points for the question according to the number and relative importance of the rubric points. This labeling scheme comprises four categories, as shown in Table~\ref{tab:status_labels}, indicating the model's response status for each rubric point. 

Recent empirical work has validated LLM-based grading with detailed rubrics, achieving correlations of r = 0.78–0.93 with human expert grading across diverse legal essay questions \cite{frankenreiter2024grading}, supporting our decomposed scoring approach. Additionally, we did not require the model to compute the total score; this calculation was instead handled during post-processing.

\begin{table}[h]
\caption{Evaluation status labels and scoring rules for OEQ rubric points.}
\label{tab:status_labels}
\begin{tabular}{lll}
\toprule
\textbf{Status} & \textbf{Description} & \textbf{Scoring Rule} \\
\midrule
\texttt{correct} & Fully correct & 70--100\% of points \\
\texttt{partial} & Partially correct & 30--70\% of points \\
\texttt{wrong} & Incorrect answer & 0--30\% of points \\
\texttt{miss} & Not covered / omitted & 0 points \\
\bottomrule
\end{tabular}
\end{table}

\subsubsection{LJP} 

Given the facts of a criminal case, we required models to generate the applicable statutory provisions, the reasoning for the judgment, and the judgment holding. Due to the lack of a standardized writing template for criminal judgments in Taiwan, coupled with their high complexity and substantial length, we defer the evaluation of judgment reasoning to future work and focus here on assessing the applicable statutory provisions and judgment holdings, which are less susceptible to statistical bias.

For evaluating judgment holdings, we considered both text generation quality metrics and sentencing accuracy. For generation quality, we employed ROUGE-1/2/L and token-level F1 scores using Jieba\footnote{\url{https://github.com/fxsjy/jieba}} for Chinese word segmentation. For sentencing accuracy, we computed verdict accuracy and sentencing precision.

For sentencing precision and accuracy, we adopted a category-specific evaluation strategy to assess the accuracy of numerical predictions. For imprisonment terms, considering the non-linear perception of sentence duration (i.e., the tolerance for error differs between long and short sentences), we followed LawBench \cite{fei-etal-2024-lawbench}, which employed a normalized log-distance metric on a logarithmic scale, with 216 months as the normalization boundary, yielding a more discriminative score ranging from 0 to 1. For detention, fines, and probation, we reverted to straightforward numerical comparisons, evaluating prediction accuracy and mean deviation within absolute thresholds (e.g., ±7 days for detention, ±1 year for probation) or relative proportions (e.g., ±10\% or ±50\% for fines). Through these metrics, we can precisely quantify the model's proficiency in predicting various types of judicial sanctions.

Finally, for the applicable statutory provisions, we considered statute accuracy, Type I error (citing non-existent/hallucinated statutes), Type II error (citing real but inapplicable statutes), and total error rate (Type I + Type II).
We employed regular expressions to extract the relevant statutes from the original judgment documents, capturing the hierarchical structure of legal provisions including Article (條), Paragraph (項), Subparagraph (款), and Item (目). The same extraction procedure was then applied to the model-generated outputs, after which the four aforementioned metrics were computed.

\section{Results}

\subsection{Overall Performance for MCQs}

Table~\ref{tab:accuracy-full-split} presents the performance of all models across 6 legal domains. We observed that nearly all models achieved their best performance on constitutional law and questions without labeled legal categories, with weaker performance on administrative law. These findings align with our expectations regarding legal question characteristics: many unlabeled questions involve entry-level examinations that primarily assess general legal concepts rather than applying the rules derived from specific statutes. Constitutional law, compared to other domains, exhibits relatively low variability and comprises fewer provisions, which contributes to stronger model performance. In contrast, administrative law encompasses a voluminous body of obscure statutes and undergoes frequent amendments, posing a greater challenge for both closed-source and open-source models.

\begin{table*}[t]
\caption{MCQs accuracy (\%) across legal domains, separated by model availability. Models are sorted by Total score within their category. Best scores in each category are \textbf{bolded}.}
\label{tab:accuracy-full-split}
\centering
\begin{tabular}{llrrrrrrr}
\toprule
\textbf{Model} & \textbf{Release} & \textbf{Const.} & \textbf{Crim.} & \textbf{Civil} & \textbf{Admin.} & \textbf{Intl.} & \textbf{No Law} & \textbf{Total} \\
\midrule
\multicolumn{9}{c}{\textit{Closed-Source Models}} \\
\midrule
claude-sonnet-4.5 & 2025-09 & \textbf{89.4} & \textbf{82.6} & \textbf{80.9} & \textbf{78.1} & 75.8 & 88.6 & \textbf{81.0} \\
gpt-5 & 2025-08 & 87.6 & 78.7 & 77.7 & 72.3 & 75.8 & \textbf{88.9} & 76.7 \\
gpt-5-2 & 2025-12 & 84.8 & 76.2 & 74.0 & 69.7 & \textbf{76.6} & 86.4 & 73.9 \\
gpt-4o & 2024-08 & 79.9 & 68.6 & 64.9 & 61.5 & 67.7 & 81.5 & 66.2 \\
\midrule
\multicolumn{9}{c}{\textit{Open-Source Models}} \\
\midrule
qwen3-235b & 2025-04 & \textbf{78.5} & \textbf{75.3} & \textbf{69.1} & \textbf{64.4} & \textbf{69.4} & \textbf{82.2} & \textbf{69.1} \\
llama-taiwan-70b-instruct & 2024-04 & 71.5 & 65.2 & 60.2 & 61.0 & 64.5 & 74.4 & 63.4 \\
llama-3.1-405b & 2024-07 & 72.2 & 63.9 & 57.6 & 56.6 & 67.7 & 73.1 & 60.5 \\
gpt-oss-120b & 2025-08 & 67.2 & 58.5 & 52.7 & 52.1 & 62.1 & 73.0 & 56.0 \\
nemotron-3-nano-30b-a3b & 2025-12 & 63.5 & 57.2 & 49.2 & 48.7 & 56.5 & 66.3 & 52.6 \\
qwen2.5-7b & 2024-09 & 59.3 & 52.1 & 46.8 & 46.7 & 53.2 & 61.2 & 49.7 \\
gpt-oss-20b & 2025-08 & 59.1 & 51.9 & 44.4 & 44.9 & 55.6 & 60.5 & 48.3 \\
llama-taiwan-8b-128k & 2024-04 & 53.6 & 47.0 & 40.5 & 42.6 & 41.1 & 57.1 & 44.9 \\
breeze-7b-instruct & 2024-03 & 46.2 & 41.6 & 35.9 & 37.4 & 46.0 & 50.5 & 39.6 \\
\bottomrule
\end{tabular}
\end{table*}

\subsection{Overall Performance for OEQs}

We report the counts of status labels in Table~\ref{tab:essay-coverage-counts}. Our findings reveal the following observations: (1) There exists a substantial performance gap in accuracy between tasks, with differentiation among models being far more pronounced in OEQs than in MCQs. For instance, gpt-4o and qwen3-235b differ by less than 3\% in total score on MCQs and exhibit closely comparable performance across all subject areas. However, on OEQs, qwen3-235b achieves more than twice the accuracy count of gpt-4o. (2) Models primarily trained on Simplified Chinese or Traditional Chinese data\footnote{This refers to stages beyond pretraining, such as the use of a higher proportion of Chinese data during SFT or RL.} can outperform larger, general-purpose models on evaluations dominated by partial and wrong labels. For example, breeze-7b-instruct outperforms gpt-oss-20b, llama-taiwan-70b-instruct outperforms gpt-oss-120b, and qwen3-235b even surpasses gpt-4o, underscoring the importance of Chinese-language corpora. (3) Due to our use of greedy decoding, llama-taiwan-8b frequently exhibited reasoning failures during generation, such as producing repetitive answers. We hypothesize that this checkpoint may have received insufficient training on reasoning-oriented texts during the SFT stage. While this limitation was not apparent in MCQs, it was substantially amplified in OEQs.

\begin{table}[h]
\caption{Number of evaluation status labels on OEQs. Models are sorted by Correct.}
\label{tab:essay-coverage-counts}
\centering
\resizebox{0.85\columnwidth}{!}{
\begin{tabular}{@{}l@{\hspace{4pt}}r@{\hspace{4pt}}r@{\hspace{4pt}}r@{\hspace{4pt}}r@{}}
\toprule
\textbf{Model} & \textbf{Correct} & \textbf{Partial} & \textbf{Wrong} & \textbf{Miss} \\
\midrule
gpt-5 & 166 & 187 & 42 & 185 \\
gpt-5-2 & 160 & 195 & 42 & 183 \\
claude-sonnet-4.5 & 160 & 176 & 59 & 185 \\
qwen3-235b & 115 & 204 & 88 & 173 \\
gpt-4o & 52 & 214 & 94 & 220 \\
llama-taiwan-70b-instruct & 55 & 184 & 104 & 237 \\
gpt-oss-120b & 53 & 185 & 126 & 216 \\
nemotron-3-nano-30b-a3b & 39 & 163 & 106 & 272 \\
llama-3.1-405b & 21 & 171 & 133 & 255 \\
qwen2.5-7b & 17 & 160 & 147 & 256 \\
breeze-7b-instruct & 10 & 124 & 168 & 278 \\
gpt-oss-20b & 17 & 132 & 92 & 339 \\
llama-taiwan-8b-128k & 5 & 50 & 55 & 470 \\
\bottomrule
\end{tabular}}
\end{table} %
\begin{figure*}[t]
  \centering
  \includegraphics[width=\linewidth]{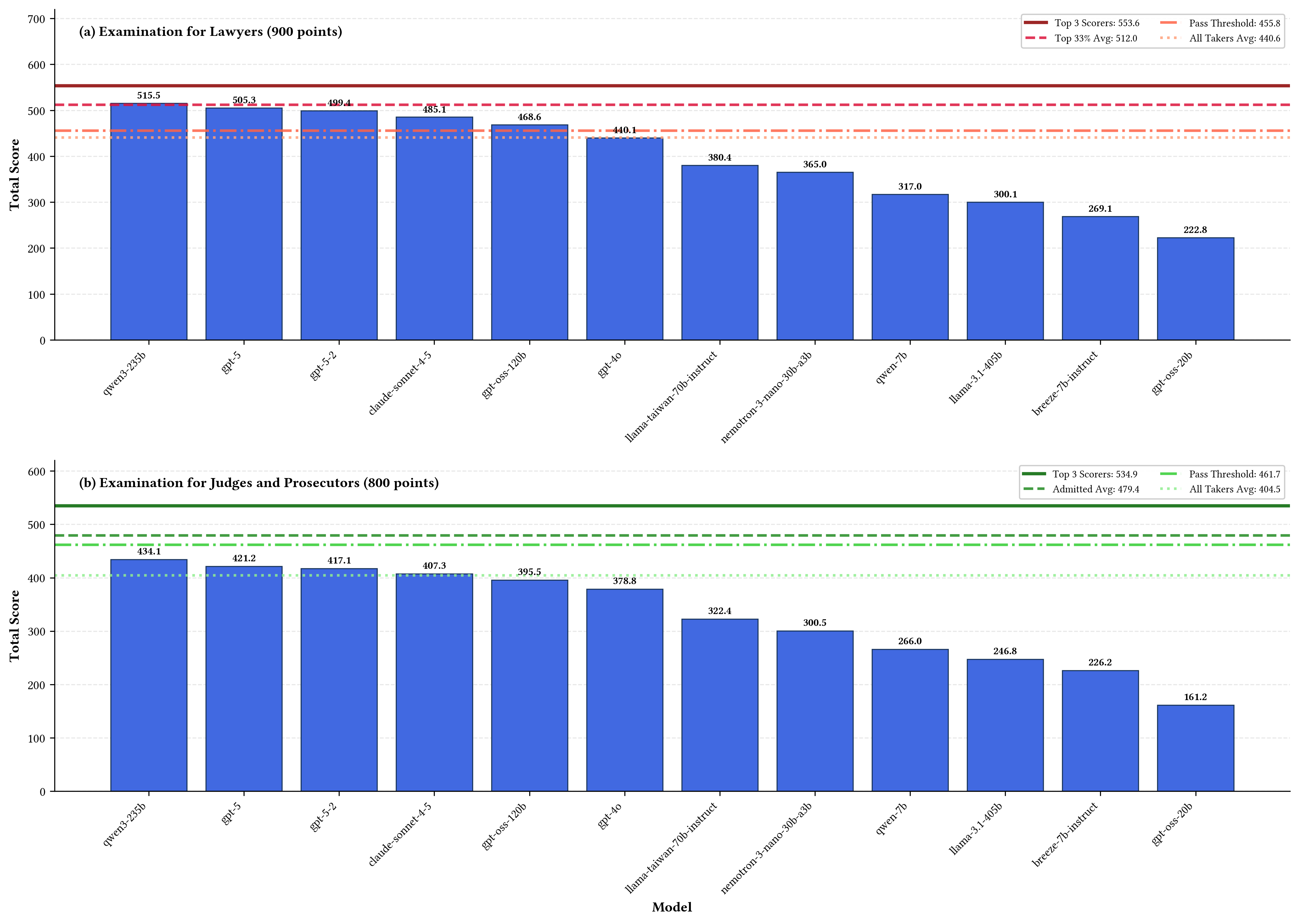}
  \caption{Model performance on OEQs compared to human examinees averaged over 2021-2024. (a) shows results for the Examination for Lawyers (900 points), and (b) shows results for the Examination for Judges and Prosecutors (800 points). Horizontal reference lines represent human benchmark statistics with Chinese essay scores deducted according to each tier (e.g., Admitted Avg scorers' Chinese essay scores for Admitted Avg scorers).}
  \label{fig:model_vs_human}
\end{figure*}

Next, we aggregated the scores output by the judge models according to the Bar Examination and the Judicial/Prosecutor Examination. To ensure fairness, we made two adjustments. First, since detailed statistics were available only for 2021–2024, we computed the model performance statistics using the four-year average, whereas the status label counts reported earlier were calculated using the complete dataset spanning all five years. Second, because both examinations include a Chinese Essay component that we did not evaluate, we could not directly compare model scores against scores from human exam-takers. Therefore, we report scores as "the score at each percentile rank minus the Chinese Essay-Writing score at the corresponding percentile rank." For example, we subtracted the average Chinese Essay score of examinees in the top 33rd percentile from their average total score to obtain the average total legal score for that percentile group.

To pass the Bar Examination, candidates are required to rank in the top 33rd percentile and achieve a total score exceeding 400 points (this score includes the Chinese Essay-Writing; when considering only legal subjects, the threshold is approximately 350 points). For the Judicial/Prosecutor Examination, the passing threshold is set at 1.2x of the number of the final recruited persons, corresponding to approximately the top 5–10\% of all examinees that sit for the second-stage exams.\footnote{The Judicial/Prosecutor Examination also requires an interview component.} These two examinations share the same exam questions but differ in the weights on subjects counted toward the final score. Figure~\ref{fig:model_vs_human} presents a comparison between model-generated results and human performance, where the model scores represent the average across both judge models. We observed that qwen3-235b achieved the best performance in both examinations, outperforming all closed-source models. Five models met the admission threshold for the Bar Examination, among which qwen3-235b, gpt-5, and gpt-5.2 approached the average score of admitted candidates. However, no model reached the passing threshold for the Judicial/Prosecutor Examination. The scores of the top 3\% of examinees exceeded that of qwen3-235b by more than 100 points, revealing a substantial performance gap.

\subsection{Overall Performance for LJP}

Table~\ref{tab:comprehensive-metrics-split-bold} presents the evaluation results for legal judgment prediction. We report the following findings: 

(1) Claude-sonnet-4.5 achieved the highest scores across nearly all metrics, particularly in sentencing prediction. (2) Nearly all models demonstrated a lack of understanding of criminal sanctions under Taiwan's Criminal Code, resulting in their inability to produce accurate sentencing predictions within the thresholds defined in our evaluation framework for one or more sentencing metrics. (3) In contrast, breeze-7b-instruct and llama-taiwan-8b achieved notably high scores. However, given their performance on OEQs and MCQs, we reasonably infer that their strong results stem from their training data containing a substantial proportion of court judgments, enabling them to perform sentencing through memorization rather than genuine legal reasoning.

\begin{table*}[t]
\caption{Comprehensive performance for main result prediction on LJP. Models are sorted by ROUGE-L.}
\label{tab:comprehensive-metrics-split-bold}
\centering
\resizebox{0.9\textwidth}{!}{
\begin{tabular}{lccccccccc}
\toprule
\textbf{Model} & \textbf{R-1} & \textbf{R-2} & \textbf{R-L} & \textbf{T-F1} & \textbf{Pri.$\pm$3m} & \textbf{Norm Log-Dist} & \textbf{Det.$\pm$7d} & \textbf{Prob.$\pm$1y} & \textbf{Fine$\pm$50\%} \\
\midrule
\multicolumn{10}{c}{\textit{Closed-Source Models}} \\
\midrule
claude-sonnet-4.5       & \textbf{0.563} & \textbf{0.412} & \textbf{0.535} & \textbf{0.564} & \textbf{0.503} & \textbf{0.907} & \textbf{0.214} & 0.146 & \textbf{0.265} \\
gpt-5-2                 & 0.462 & 0.303 & 0.427 & 0.463 & 0.394 & 0.876 & 0.129 & 0.219 & 0.000 \\
gpt-5                   & 0.432 & 0.267 & 0.399 & 0.433 & 0.404 & 0.872 & 0.057 & 0.219 & 0.000 \\
gpt-4o                  & 0.306 & 0.160 & 0.284 & 0.306 & 0.165 & 0.834 & 0.000 & \textbf{0.274} & 0.042 \\
\midrule
\multicolumn{10}{c}{\textit{Taiwan Open-Source Models}} \\
\midrule
llama-taiwan-8b-128k         & \textbf{0.414} & 0.260 & \textbf{0.387} & \textbf{0.417} & \textbf{0.568} & \textbf{0.898} & \textbf{0.103} & \textbf{0.604} & \textbf{0.102} \\
breeze-7b-instruct      & 0.399 & \textbf{0.279} & 0.383 & 0.399 & 0.319 & 0.886 & 0.000 & 0.106 & 0.000 \\
\midrule
\multicolumn{10}{c}{\textit{Other Open-Source Models}} \\
\midrule
qwen3-235b              & \textbf{0.358} & \textbf{0.209} & \textbf{0.325} & \textbf{0.358} & \textbf{0.215} & \textbf{0.846} & \textbf{0.029} & \textbf{0.500} & 0.000 \\
llama-3.1-405b          & 0.319 & 0.169 & 0.298 & 0.319 & 0.181 & 0.829 & 0.000 & 0.219 & 0.000 \\
gpt-oss-120b            & 0.300 & 0.138 & 0.263 & 0.302 & 0.043 & 0.753 & 0.000 & 0.208 & \textbf{0.041} \\
gpt-oss-20b             & 0.262 & 0.130 & 0.246 & 0.262 & 0.013 & 0.704 & 0.000 & 0.000 & 0.000 \\
nemotron-3-nano-30b-a3b & 0.239 & 0.115 & 0.225 & 0.240 & 0.065 & 0.741 & 0.000 & 0.053 & 0.000 \\
qwen2.5-7b                 & 0.235 & 0.107 & 0.217 & 0.235 & 0.067 & 0.777 & 0.000 & 0.125 & 0.000 \\
\bottomrule
\end{tabular}}
\end{table*}

We further examine the error rates for applicable statutory provisions in Table~\ref{tab:hallucination-rates-pct}, several findings emerge: (1) Although claude-sonnet-4.5 achieved the best performance among closed-source models, its Type I error rate (0.2406) was relatively high, indicating a tendency to hallucinate statutes that do not exist. \\(2) While gpt-oss-120b exhibited an extremely low accuracy rate (0.0031), it achieved the lowest Type I (non-existent statute) error rate among all models (0.0992). This suggests that although its predictions were largely incorrect, the errors predominantly involved citing irrelevant but existing statutes (Type II error as high as 0.8976), rather than fabricating statutes entirely. (3) llama-taiwan-8b achieved the highest accuracy at 0.0993 and the lowest total error rate (0.9007). Consistent with our earlier conclusions, this demonstrates that models trained on Taiwan-specific corpora exhibit significantly higher accuracy in statutory citation compared to other models, which we attribute to memorization effects.

\begin{table}[!htbp]
\caption{Statute citation hallucination rates (\%). Best scores in each category are \textbf{bolded} (Highest for Acc; Lowest for errors).}
\label{tab:hallucination-rates-pct}
\centering
\setlength{\tabcolsep}{3.5pt} %
\begin{tabular}{lcccc}
\toprule
\textbf{Model} & \textbf{Acc} & \textbf{TypeI} & \textbf{TypeII} & \textbf{Total} \\
\midrule
\multicolumn{5}{c}{\textit{Closed-Source Models}} \\
\midrule
claude-sonnet-4.5 & \textbf{5.7} & 24.1 & \textbf{70.2} & \textbf{94.3} \\
gpt-5-2 & 3.9 & \textbf{12.3} & 83.8 & 96.1 \\
gpt-5 & 2.6 & 15.5 & 81.8 & 97.4 \\
gpt-4o & 2.0 & 24.3 & 73.7 & 98.0 \\
\midrule
\multicolumn{5}{c}{\textit{Taiwan Models}} \\
\midrule
llama-taiwan-8b-128k & \textbf{9.9} & 15.1 & \textbf{75.0} & \textbf{90.1} \\
breeze-7b-instruct & 2.6 & \textbf{13.8} & 83.6 & 97.4 \\
\midrule
\multicolumn{5}{c}{\textit{Other Open-Source Models}} \\
\midrule
qwen3-235b & \textbf{2.4} & 16.8 & 80.7 & \textbf{97.6} \\
llama-3.1-405b & 1.1 & 12.2 & 86.7 & 98.9 \\
qwen2.5-7b & 0.6 & 15.8 & 83.6 & 99.4 \\
gpt-oss-20b & 0.4 & 28.3 & \textbf{71.3} & 99.6 \\
nemotron-3-nano-30b & 0.3 & 18.7 & 81.0 & 99.7 \\
gpt-oss-120b & 0.3 & \textbf{9.9} & 89.8 & 99.7 \\
\bottomrule
\end{tabular}
\end{table}

\section{Discussion}

\subsection{Extreme Hard Questions for MCQs}

We analyzed the MCQs that nearly all models answered incorrectly. Among the 16,493 questions, 290 (1.76\%) stumped all models, and 550 (3.33\%) were answered correctly by fewer than 10\% of models. Among the questions that all models failed, civil law appeared most frequently but accounted for only 11 questions, while only three legal categories contained more than 10 questions. Remarkably, the 290 questions spanned 149 unique statutes, the vast majority of which pertain to complex and less-known regulations, such as the National Palace Museum Organization Act (國立故宮博物院組織法) or the 37.5\% Arable Rent Reduction Act (耕地三七五減租條例), which are highly specific to civil service examinations.

The Civil Code questions that stumped all models predominantly involve estate distribution or statute of limitations, which require preliminary calculations to solve. Figure~\ref{fig:civil-law-failure} illustrates a civil law question on estate distribution. This question pertains to Art. 1173 of the Civil Code, which addresses the concept of investments being classified as special gifts and treated as advancement of inheritance (歸扣). Among the 13 models evaluated, 11 selected option B, while 2 selected option A.

\subsection{Language and Terminological Effects}

To investigate whether language and terminology influence model performance, we translated the 2024 MCQs into English and Simplified Chinese using claude-sonnet-4.5. The English translations were guided by the "Commonly Used Legal Vocabularies in Courts and Litigation Procedures" published by the Judicial Yuan.\footnote{\url{https://www.judicial.gov.tw/tw/cp-1778-90025-35329-1.html}} We report the results for gpt-4o. Although the overall average accuracy remained similar across languages, notable variations were observed across individual legal domains.

Performance improved on civil service-related statutes when questions were translated into English, including the Act on Property-Declaration by Public Servants (公職人員財產申報法) and the Fair Trade Act. We attribute this improvement to the legal origins of these statutes: Taiwan's Act on Property-Declaration by Public Servants draws heavily from Sunshine Laws and the Ethics in Government Act of 1978 of the USA, yielding precise English-Chinese terminological correspondence. Conversely, performance declined on areas involving jurisdiction-specific numerical thresholds, such as the People with Disabilities Rights Protection Act, the Settlement of Labor Disputes Act, tax codes, and border control regulations.

Claude-sonnet-4.5 tended to paraphrase formal legal terminology into more colloquial expressions for the Simplified Chinese translations. Such simplification improved comprehension for certain statutes, such as the Estate and Gift Tax Act. However, for Taiwan-specific regulations containing domain-specific terminology that resists simplification--such as the Regulations Governing Assessment of Profit-Seeking Enterprise Income Tax (營利事業所得稅查核準則) and the Equalization of Land Rights Act (平均地權條例)--performance declined markedly. A systematic comparative analysis of legal sources and terminological conventions across different legal systems is deferred to future work.

\subsection{Time Insensitivity and Data Leakage}

Table~\ref{tab:accuracy-by-year} presents the performance of all models on MCQs by examination year. We observed that model performance on MCQs does not exhibit temporal sensitivity, even when a given checkpoint was trained prior to the release of the examination questions. We hypothesize that this is attributable to examination policies. Examination committees generally avoid questions involving controversial or highly time-sensitive topics, or instead permit multiple correct answers for such questions, which are excluded from our data preprocessing stage, to avoid penalizing examinees for insufficient familiarity with recently amended statutes.

\begin{table}[t]
\caption{MCQs accuracy across examination years. Models are sorted by the score for 2024 questions. }
\label{tab:accuracy-by-year}
\begin{tabular}{@{}l@{\hspace{4pt}}r@{\hspace{4pt}}r@{\hspace{4pt}}r@{\hspace{4pt}}r@{\hspace{4pt}}r@{}}
\toprule
\textbf{Model} & \textbf{2020} & \textbf{2021} & \textbf{2022} & \textbf{2023} & \textbf{2024} \\
\midrule
claude-sonnet-4.5         & 81.3 & 81.0 & 80.1 & 82.1 & 80.1 \\
gpt-5                     & 75.9 & 77.6 & 75.8 & 77.5 & 76.7 \\
gpt-5-2                   & 74.2 & 73.4 & 72.8 & 74.3 & 74.9 \\
qwen3-235b                & 68.5 & 69.5 & 68.1 & 69.4 & 70.5 \\
gpt-4o                    & 65.7 & 66.4 & 65.9 & 67.1 & 66.1 \\
llama-taiwan-70b-instruct & 64.6 & 61.8 & 63.9 & 64.0 & 62.3 \\
llama-3.1-405b            & 60.5 & 59.8 & 59.9 & 61.4 & 60.7 \\
gpt-oss-120b              & 56.9 & 55.6 & 55.8 & 55.4 & 56.2 \\
nemotron-3-nano-30b-a3b   & 51.2 & 52.8 & 51.6 & 53.9 & 53.9 \\
gpt-oss-20b               & 47.0 & 48.3 & 47.5 & 48.6 & 50.4 \\
qwen2.5-7b                & 49.7 & 50.1 & 48.9 & 50.4 & 49.4 \\
llama-taiwan-8b-128k      & 44.5 & 44.9 & 45.2 & 44.5 & 45.8 \\
breeze-7b-instruct        & 39.8 & 39.3 & 39.9 & 39.0 & 39.8 \\
\bottomrule
\end{tabular}
\end{table}

Similarly, the table allows us to assess the potential issue of data leakage, as these examination questions are publicly discussed on online platforms and may have been collected into the training corpora of these LLMs. If such contamination had occurred, we would expect substantially higher accuracy rates on earlier questions before the training cutoffs and lower accuracy for later questions. However, we find that performance on the 2024 examinations does not differ significantly from that of other years.

\section{Limitations}
While our evaluation uses official Ministry of Examination rubrics for OEQs, and recent work shows LLM-based grading can correlate highly with human experts \cite{frankenreiter2024grading}, we essentially compared AI-generated answers assessed by an AI judge against the human-graders' assessment for human exam-takers. Therefore, the primary value of our OEQ evaluation lies in understanding whether models accurately address the reasoning processes that examination committees consider important during deep reasoning tasks, rather than in determining whether AI models could outcompete human law-school graduates before human-graders. 

In addition, data leakage remains an unavoidable concern for LJP, as judgment data has long been considered a valuable open-source dataset in Traditional Chinese and Taiwanese culture. Although legal knowledge memorization is an important task in civil law benchmarks such as LawBench \cite{fei-etal-2024-lawbench}, we recommend against evaluating LLMs' memorization ability, since data leakage would likely yield overly optimistic results.

\section{Conclusion}

We present TW-LegalBench, a benchmark for evaluating LLMs on legal reasoning in Taiwanese Mandarin, comprising over 16,000 multiple-choice questions with statute-level annotations spanning 43 law types, 117 open-ended essay questions from professional examinations with official scoring rubrics, and 14,000+ legal judgment prediction instances covering 107 crime categories.

Our evaluation of 13 LLMs reveals several findings. First, top-performing models such as claude-sonnet-4.5 and qwen3-235b exceed the admission threshold for the Bar Examination, yet all models fall short of the cutoff for judges and prosecutors, indicating a substantial gap between models and human experts. Second, models demonstrate reasonable performance on verdict type prediction and sentencing estimation but struggle with statutory citation; the best-performing models achieve less than 10\% accuracy. Third, models trained on more Traditional Chinese or Taiwan-specific legal corpora consistently outperform larger general-purpose models on open-ended tasks and judgment prediction, underscoring the importance of jurisdiction-specific training data.

\section{Acknowledgement}
We are grateful to Mr. Zhi Rui Tam for the insightful discussions that contributed to this work. This research was supported in part by the computational resources provided by the Behavioral and Data Science Research Center, National Taiwan University.

\bibliographystyle{ACM-Reference-Format}
\bibliography{sample-base}

@article{hendryckstest2021,
  title={Measuring Massive Multitask Language Understanding},
  author={Dan Hendrycks and Collin Burns and Steven Basart and Andy Zou and Mantas Mazeika and Dawn Song and Jacob Steinhardt},
  journal={Proceedings of the International Conference on Learning Representations (ICLR)},
  year={2021}
}

@article{llama3herd,
  title={The Llama 3 Herd of Models},
  author={{Meta AI}},
  journal={arXiv preprint arXiv:2407.21783},
  year={2024},
  url={https://arxiv.org/abs/2407.21783}
}

@article{Yang2024Qwen25TR,
  title={Qwen2.5 Technical Report},
  author={Qwen Team},
  journal={ArXiv},
  year={2024},
  volume={abs/2412.15115},
  url={https://api.semanticscholar.org/CorpusID:274859421}
}

@misc{yang2025qwen3technicalreport,
      title={Qwen3 Technical Report}, 
      author={Qwen Team},
      year={2025},
      eprint={2505.09388},
      archivePrefix={arXiv},
      primaryClass={cs.CL},
      url={https://arxiv.org/abs/2505.09388}, 
}

@misc{OpenAI_GPT4_2023,
  author       = {{OpenAI}},
  title        = {{GPT-4 Technical Report}},
  year         = {2023},
  eprint       = {2303.08774},
  archivePrefix = {arXiv},
  primaryClass = {cs.CL},
  }

@misc{OpenAI2025GPT5,
      title={OpenAI GPT-5 System Card}, 
      author={OpenAI},
      year={2025},
      eprint={2601.03267},
      archivePrefix={arXiv},
      primaryClass={cs.CL},
      url={https://arxiv.org/abs/2601.03267}, 
}

@article{DBLP:journals/corr/abs-2403-20180,
  author       = {Po{-}Heng Chen and
                  Sijia Cheng and
                  Wei{-}Lin Chen and
                  Yen{-}Ting Lin and
                  Yun{-}Nung Chen},
  title        = {Measuring Taiwanese Mandarin Language Understanding},
  journal      = {CoRR},
  volume       = {abs/2403.20180},
  year         = {2024},
  url          = {https://doi.org/10.48550/arXiv.2403.20180},
  doi          = {10.48550/ARXIV.2403.20180},
  eprinttype    = {arXiv},
  eprint       = {2403.20180},
  bibsource    = {dblp computer science bibliography, https://dblp.org}
}

@inproceedings{
tam2024tmmlu,
title={{TMMLU}+: An Improved Traditional Chinese Evaluation Suite for Foundation Models},
author={Zhi Rui Tam and Ya Ting Pai and Yen-Wei Lee and Hong-Han Shuai and Jun-Da Chen and Wei Min Chu and Sega Cheng},
booktitle={First Conference on Language Modeling},
year={2024},
url={https://openreview.net/forum?id=95TayIeqJ4}
}

@inproceedings{10.5555/3666122.3668037,
author = {Guha, Neel and Nyarko, Julian and Ho, Daniel E. and R\'{e}, Christopher and Chilton, Adam and Narayana, Aditya and Chohlas-Wood, Alex and Peters, Austin and Waldon, Brandon and Rockmore, Daniel N. and Zambrano, Diego and Talisman, Dmitry and Hoque, Enam and Surani, Faiz and Fagan, Frank and Sarfaty, Galit and Dickinson, Gregory M. and Porat, Haggai and Hegland, Jason and Wu, Jessica and Nudell, Joe and Niklaus, Joel and Nay, John and Choi, Jonathan H. and Tobia, Kevin and Hagan, Margaret and Ma, Megan and Livermore, Michael and Rasumov-Rahe, Nikon and Holzenberger, Nils and Kolt, Noam and Henderson, Peter and Rehaag, Sean and Goel, Sharad and Gao, Shang and Williams, Spencer and Gandhi, Sunny and Zur, Tom and Iyer, Varun and Li, Zehua},
title = {LEGALBENCH: a collaboratively built benchmark for measuring legal reasoning in large language models},
year = {2023},
publisher = {Curran Associates Inc.},
address = {Red Hook, NY, USA},
booktitle = {Proceedings of the 37th International Conference on Neural Information Processing Systems},
articleno = {1915},
numpages = {157},
location = {New Orleans, LA, USA},
series = {NIPS '23}
}

@inproceedings{fei-etal-2024-lawbench,
    title = "{L}aw{B}ench: Benchmarking Legal Knowledge of Large Language Models",
    author = "Fei, Zhiwei  and
      Shen, Xiaoyu  and
      Zhu, Dawei  and
      Zhou, Fengzhe  and
      Han, Zhuo  and
      Huang, Alan  and
      Zhang, Songyang  and
      Chen, Kai  and
      Yin, Zhixin  and
      Shen, Zongwen  and
      Ge, Jidong  and
      Ng, Vincent",
    editor = "Al-Onaizan, Yaser  and
      Bansal, Mohit  and
      Chen, Yun-Nung",
    booktitle = "Proceedings of the 2024 Conference on Empirical Methods in Natural Language Processing",
    month = nov,
    year = "2024",
    address = "Miami, Florida, USA",
    publisher = "Association for Computational Linguistics",
    url = "https://aclanthology.org/2024.emnlp-main.452/",
    doi = "10.18653/v1/2024.emnlp-main.452",
    pages = "7933--7962"
}

@article{GU2026101253,
title = {A survey on LLM-as-a-Judge},
journal = {The Innovation},
pages = {101253},
year = {2026},
issn = {2666-6758},
doi = {https://doi.org/10.1016/j.xinn.2025.101253},
url = {https://www.sciencedirect.com/science/article/pii/S2666675825004564},
author = {Jiawei Gu and Xuhui Jiang and Zhichao Shi and Hexiang Tan and Xuehao Zhai and Chengjin Xu and Wei Li and Yinghan Shen and Shengjie Ma and Honghao Liu and Saizhuo Wang and Kun Zhang and Zhouchi Lin and Bowen Zhang and Lionel Ni and Wen Gao and Yuanzhuo Wang and Jian Guo}
}

@inproceedings{
han2025lawshift,
title={LawShift: Benchmarking Legal Judgment Prediction Under Statute Shifts},
author={Zhuo Han and Yi Yang and Yi Feng and Wanhong Huang and Xuxing Ding and Chuanyi Li and Jidong Ge and Vincent Ng},
booktitle={The Thirty-ninth Annual Conference on Neural Information Processing Systems Datasets and Benchmarks Track},
year={2025},
url={https://openreview.net/forum?id=5SpFenlxDF}
}

@inproceedings{singh-etal-2025-global,
    title = "Global {MMLU}: Understanding and Addressing Cultural and Linguistic Biases in Multilingual Evaluation",
    author = "Singh, Shivalika  and
      Romanou, Angelika  and
      Fourrier, Cl{\'e}mentine  and
      Adelani, David Ifeoluwa  and
      Ngui, Jian Gang  and
      Vila-Suero, Daniel  and
      Limkonchotiwat, Peerat  and
      Marchisio, Kelly  and
      Leong, Wei Qi  and
      Susanto, Yosephine  and
      Ng, Raymond  and
      Longpre, Shayne  and
      Ruder, Sebastian  and
      Ko, Wei-Yin  and
      Bosselut, Antoine  and
      Oh, Alice  and
      Martins, Andre  and
      Choshen, Leshem  and
      Ippolito, Daphne  and
      Ferrante, Enzo  and
      Fadaee, Marzieh  and
      Ermis, Beyza  and
      Hooker, Sara",
    editor = "Che, Wanxiang  and
      Nabende, Joyce  and
      Shutova, Ekaterina  and
      Pilehvar, Mohammad Taher",
    booktitle = "Proceedings of the 63rd Annual Meeting of the Association for Computational Linguistics (Volume 1: Long Papers)",
    month = jul,
    year = "2025",
    address = "Vienna, Austria",
    publisher = "Association for Computational Linguistics",
    url = "https://aclanthology.org/2025.acl-long.919/",
    doi = "10.18653/v1/2025.acl-long.919",
    pages = "18761--18799",
    ISBN = "979-8-89176-251-0"
}

@inproceedings{chen2025continual,
  title     = {Continual Pre-Training is (not) What You Need in Domain Adaptation},
  author    = {Chen, Pin-Er and Lian, Da-Chen and Chi, Jou-An and Hsieh, Shu-Kai and Huang, Sieh-Chuen and Shao, Hsuan-Lei and Chiu, Jun-Wei and Lin, Yang-Hsien and Chen, Zih-Ching and Lee, Cheng-Kuang and Huang, Eddie TC and See, Simon},
  booktitle = {Proceedings of the Asian Conference on Machine Learning},
  year      = {2025},
  editor    = {Lee, Hung-yi and Liu, Tongliang},
  volume    = {304},
  series    = {Proceedings of Machine Learning Research},
  publisher = {PMLR}
}

@misc{jiang2023mistral7b,
      title={Mistral 7B}, 
      author={Albert Q. Jiang and Alexandre Sablayrolles and Arthur Mensch and Chris Bamford and Devendra Singh Chaplot and Diego de las Casas and Florian Bressand and Gianna Lengyel and Guillaume Lample and Lucile Saulnier and Lélio Renard Lavaud and Marie-Anne Lachaux and Pierre Stock and Teven Le Scao and Thibaut Lavril and Thomas Wang and Timothée Lacroix and William El Sayed},
      year={2023},
      eprint={2310.06825},
      archivePrefix={arXiv},
      primaryClass={cs.CL},
      url={https://arxiv.org/abs/2310.06825}, 
}

@misc{hsu2024breeze7btechnicalreport,
      title={Breeze-7B Technical Report}, 
      author={Chan-Jan Hsu and Chang-Le Liu and Feng-Ting Liao and Po-Chun Hsu and Yi-Chang Chen and Da-Shan Shiu},
      year={2024},
      eprint={2403.02712},
      archivePrefix={arXiv},
      primaryClass={cs.CL},
      url={https://arxiv.org/abs/2403.02712}, 
}

@misc{lin2023taiwanllmbridginglinguistic,
      title={Taiwan LLM: Bridging the Linguistic Divide with a Culturally Aligned Language Model}, 
      author={Yen-Ting Lin and Yun-Nung Chen},
      year={2023},
      eprint={2311.17487},
      archivePrefix={arXiv},
      primaryClass={cs.CL},
      url={https://arxiv.org/abs/2311.17487}, 
}

@inproceedings{10.5555/3600270.3602070,
author = {Wei, Jason and Wang, Xuezhi and Schuurmans, Dale and Bosma, Maarten and Ichter, Brian and Xia, Fei and Chi, Ed H. and Le, Quoc V. and Zhou, Denny},
title = {Chain-of-thought prompting elicits reasoning in large language models},
year = {2022},
isbn = {9781713871088},
publisher = {Curran Associates Inc.},
address = {Red Hook, NY, USA},
booktitle = {Proceedings of the 36th International Conference on Neural Information Processing Systems},
articleno = {1800},
numpages = {14},
location = {New Orleans, LA, USA},
series = {NIPS '22}
}

@misc{openai2025gptoss120bgptoss20bmodel,
      title={gpt-oss-120b \& gpt-oss-20b Model Card}, 
      author={OpenAI},
      year={2025},
      eprint={2508.10925},
      archivePrefix={arXiv},
      primaryClass={cs.CL},
      url={https://arxiv.org/abs/2508.10925}, 
}

@misc{nvidia2025nemotron3nanoopen,
  author       = {{NVIDIA}},
  title        = {{Nemotron 3 Nano: Open, Efficient Mixture-of-Experts Hybrid Mamba-Transformer Model for Agentic Reasoning}},
  year         = {2025},
  eprint       = {2512.20848},
  archivePrefix = {arXiv},
  primaryClass = {cs.CL},
  url          = {https://arxiv.org/abs/2512.20848}
}

@article{frankenreiter2024grading,
  title={Grading Machines: Can AI Exam-Grading Replace Law Professors?},
  author={Frankenreiter, Jens and Cope, Kevin L and Hirst, Scott and Posner, Eric A and Schwarcz, Daniel and Thorley, Dane},
  journal={SSRN Electronic Journal},
  year={2024},
  doi={10.2139/ssrn.5851362},
  url={https://papers.ssrn.com/sol3/papers.cfm?abstract_id=5851362}
}

\end{document}